\documentclass[10pt,twocolumn,letterpaper]{article}

\usepackage{ijcb}
\usepackage{times}
\usepackage{epsfig}
\usepackage{graphicx}
\usepackage{amsmath}
\usepackage{amssymb}
\usepackage{cite}
\usepackage{setspace}
\usepackage{enumerate}
\usepackage[norule,symbol,perpage]{footmisc}
\makeatletter
\newcommand*{\rom}[1]{\expandafter\@slowromancap\romannumeral #1@}
\makeatother

\ijcbfinalcopy 

\ifijcbfinal\fi

\makeatletter
\def\ps@IEEEtitlepagestyle{
\def\@oddfoot{\mycopyrightnotice}
\def\@evenfoot{}
}
\def\mycopyrightnotice{
{\hfill \footnotesize 978-1-7281-9186-7/20/\$31.00 \copyright 2020 IEEE\hfill}
}
\makeatother

\begin{document}

\title{Cross-Spectral Iris Matching Using Conditional Coupled GAN}

\author{Moktari Mostofa\\
\and
Fariborz Taherkhani\\
\and
Jeremy Dawson\\
\and
Nasser M. Nasrabadi\\
\and
West Virginia University\\
\tt\small{\{mm0251,ft0009\}}@mix.wvu.edu, \{jeremy.dawson,  nasser.nasrabadi\}@mail.wvu.edu}

\maketitle
\thispagestyle{empty}

\begin{abstract}
Cross-spectral iris recognition is emerging as a promising biometric approach to authenticating the identity of individuals. However, matching iris images acquired at different spectral bands shows significant performance degradation when compared to single-band near-infrared (NIR) matching due to the spectral gap between iris images obtained in the NIR and visual-light (VIS) spectra. Although researchers have recently focused on deep-learning-based approaches to recover invariant representative features for more accurate recognition performance, the existing methods cannot achieve the expected accuracy required for commercial applications. Hence, in this paper, we propose a conditional coupled generative adversarial network (CpGAN) architecture for cross-spectral iris recognition by projecting the VIS and NIR iris images into a low-dimensional embedding domain to explore the hidden relationship between them. The conditional CpGAN framework consists of a pair of GAN-based networks, one responsible for retrieving images in the visible domain and other responsible for retrieving images in the NIR domain. Both networks try to map the data into a common embedding subspace to ensure maximum pair-wise similarity between the feature vectors from the two iris modalities of the same subject. To prove the usefulness of our proposed approach, extensive experimental results obtained on the PolyU dataset are compared to existing state-of-the-art cross-spectral recognition methods.
\end{abstract}
\section{Introduction}

Iris recognition has received considerable attention in personal identification \cite{bowyer2008image, jain201650} due to highly distinctive spatial texture patterns in iris. It is considered as one of the most reliable and secure identity verification methods in biometrics \cite{li2017accurate, chen2014efficient}. The human iris pattern is observed to have unique and different textures due to the process of chaotic morphogenesis that causes its formation in early childhood, exhibiting variation even among identical twins. Therefore, iris recognition has been extensively used in ID authentication tasks. Many applications require both probe and gallery iris images to be captured in the same optical spectrum, under either near-infrared (NIR) or visual light (VIS), for homogeneous iris recognition. Recently, high-resolution visible surveillance cameras that can capture useable opportunistic iris images have enabled biometric systems that could potentially  compare these visible iris images to a NIR gallery using cross-spectral matching. 
\begin{figure}[t]
 \centering
 \includegraphics[width=9cm]{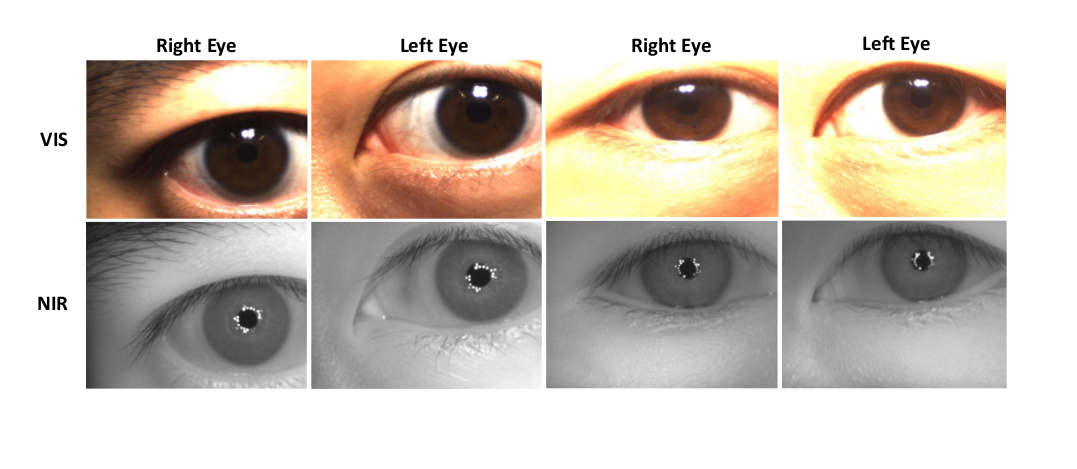}
 \caption{VIS and NIR iris images from the PolyU bi-spectral iris database.}
 \label{fig:profile_frontal_images}
\end{figure}
Cross-spectral iris matching is defined as the ability of matching iris images acquired in different spectral bands (e.g., VIS at 400-750 nm wavelength and NIR at 750-1400 nm wavelength) \cite{aguilera2017local}.

Therefore, to facilitate effective iris matching, cross-spectral iris recognition systems have recently been developed \cite{ramaiah2016matching, behera2017periocular, oktiana2019advances, nalla2016toward}. However, existing methods still suffer from significant performance degradation \cite{wang2019cross}. The spectral difference is believed to be the major reason, which yields poor recognition performance. As shown in Fig.1, the visual differences between the VIS and NIR iris images make it obvious that the choice of illumination spectrum plays a vital role in  emphasizing imaged iris patterns. For instance, iris textures are clearly visible in the VIS spectrum and complex patterns are even highlighted under the VIS illumination. However, the recognition performance is highly affected by reflection that occlude the iris pattern in certain regions. On the other hand, although almost all of the prominent iris texture patterns are missing in the NIR images, the iris recognition in the NIR images is more efficient compared to iris recognition in VIS images due to less reflections. Therefore, matching iris images in cross-spectral domain has become a challenging task, which requires to be explored to achieve a high accuracy in cross-spectral iris matching.

Previous research shows that the most essential inner properties of an image can be mapped to a reduced low-dimensional latent subspace. A Latent subspace is a compressed representation of the image space, which contains the most relevant and useful features of the raw data. In this
paper, we hypothesize that iris images in the VIS domain are connected to the iris images in the NIR domain in a low-dimensional latent embedded feature subspace. Our goal is to explore this hidden correlation by projecting VIS iris images and NIR iris images into a common latent embedding subspace. Moreover, we posit that, if we perform verification in the latent domain, matching results would be more accurate due to the shared common features in that domain. Therefore,
we propose a deep coupled learning framework for cross-spectral iris matching, which utilizes a conditional coupled generative adversarial network (CpGAN) to learn a common embedded feature vector via exploring the correlation between the NIR and VIS iris images in a reduced dimensional latent
embedding feature subspace. The key benefits from our proposed iris recognition approach can be summarized as the following:
\begin{itemize}
 \item {A novel framework for cross-spectral iris matching using coupled generative adversarial network has been proposed.}
\item{Comprehensive experiments using a benchmark PolyU Bi-Spectral dataset with comparable results against the baseline methods ascertain the validity of the proposed CpGAN framework.}

\item{The proposed framework investigates the potential capabilities of GAN based network to improve the performance of traditional cross-spectral iris recognition methods.}

\end{itemize}

\section{Literature Review}

In recent years, cross-spectral iris matching has gained significant interest in the biometric research community for security, national ID programs, and also for personal identity verification purposes \cite{ramaiah2016matching, behera2017periocular, oktiana2019advances, nalla2016toward}. The accuracy of an iris recognition system most importantly depends on the feature extraction approaches. Hence, a robust feature extraction method used for representing iris texture patterns is essential in cross-spectral iris matching. Oktiana et al. \cite{oktiana2018features} provides a description of several feature representation methods based on the VIS and NIR imaging systems. Among them, LBP and BSIF are the best feature descriptors, which have been found \cite{oktiana2018features} to accurately extract the texture patterns of the iris for cross-spectral matching. 

In \cite{vyas2019cross} the authors proposed a feature descriptor, which applies a 2D Gabor filter bank to compute the iris pattern at multiple scales and orientation. The iris images captured in the VIS spectrum often suffer from noise due to illumination occlusions and position shifting. Therefore, they utilized the difference of variance (DoV) features to divide the iris template into sub-blocks, as the DoV features are invariant to noise. However, this method could not achieve the high accuracy required for practical applications (high EER of 31.08\%) because it is unable to relate the information comprised in the NIR and VIS images.

In the work of Abdullah et al. \cite{abdullah2016novel}, the matching accuracy has increased with a 24.28\% decrease in EER. They employed a 1D log-Gabor-filter with three different descriptors, namely the Gabor difference of Gaussian (G-DoG), Gabor binarized statistical image features (G-BSIF), and Gabor multiscale weberface (G-MSW), and achieved much lower EER of 6.8\%. It is also considered as the most accurate performance of cross-spectral iris recognition method, according to the report in \cite{sequeira2016cross}. 

With the advent of convolutional neural networks (CNN), cross-spectral iris recognition research efforts have concentrated more towards feature learning through convolutional layers \cite{wang2019cross}. In \cite{wang2019cross} the authors observed that CNN-based features carry sparse information and offer a compact representation for the iris template, which is significantly reduced in size. Moreover, this approach incorporates supervised discrete hashing on the learned features to achieve excellent results compared to other CNN-based iris recognition methods. Their proposed method resulted in an EER of 5.39\%. 

\section{Generative Adversarial Network}


Recently, GANs have achieved considerable attention from the deep learning research community due to their significant contributions in image generation tasks. The basic GAN framework consists of two modules – a generator module, G, and a discriminator module, D. The objective of the generator, G, is to learn a mapping, $G:z\rightarrow y$, so that it can produce synthesized samples from a noise variable, $z$, with a prior noise distribution, $p_{z}(z)$, which is difficult for the discriminator, D, to distinguish from the real data distribution, $p_{data}$, over $y$. The generator, $G(z; \theta_{g})$ is a differentiable function which is trained with parameters $\theta_{g}$ when mapping the noise variable, $z$, to the actual data space, $y$. Simultaneously, the discriminator, D, is trained as a binary classifier with parameters $\theta_{d}$ such that it can distinguish the real samples, $y$, from the fake  ones, $G(z)$. Both the generator and discriminator networks compete with each other in a two-player minimax game. We calculate the following loss function, $L(D,G)$, for the GAN:
\vspace{-0.25cm}
\begin{equation}\begin{split}
\centering
     L(D,G) & = E_{y\sim P_{data}(y)}[\log D(y)]\\ & + E_{z\sim P_{z}(z)}[\log (1-D(G(z)))]. \end{split}
 \end{equation}

The objective function of GAN defines the term “two player minimax game” by optimizing the loss function, $L(D,G)$, as follows: 
\begin{equation}\begin{split}
     \min_{G}\max_{D} L(D,G) & =\min_{G}\max_{D}[E_{y\sim P_{data}(y)}[\log D(y)]\\ & + E_{z\sim P_{z}(z)}[\log (1-D(G(z)))]].\end{split}\label{eq:2}
\end{equation}

\begin{figure*}
\centering
\includegraphics[width=11.5 cm]{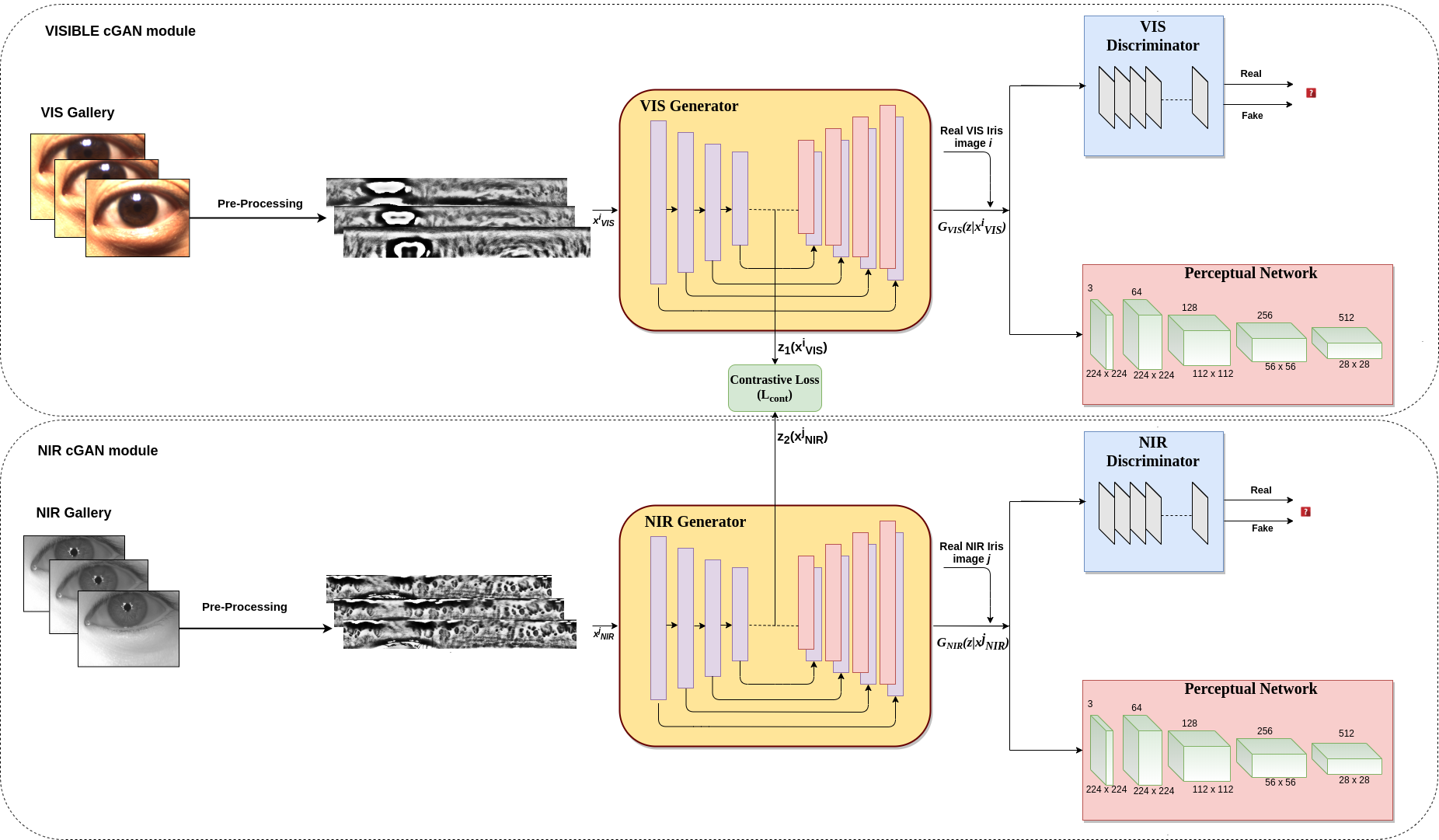}
\caption{Architecture of our proposed conditional CpGAN framework. During training, the contrastive loss function is used in the latent embedding subspace to optimize the network parameters so that latent features of iris images from different spectral domain of the same identity are close to each other while the features of different identities are pushed further apart.}\label{fig:arch}
\vspace{-5.5mm}
\vspace{0.8cm}
\end{figure*}

One of the variants of GAN is introduced in \cite{mirza_2014_conditional} as the conditional GAN (cGAN) which expands the scope of synthesized image generation by setting a condition for both the generative and discriminative networks. The cGAN applies an auxiliary variable, $x$, as a condition which could be any kind of useful information such as texts \cite{RAYLLS_16}, images \cite{Isola_2017_ImagetoImageTW} or discrete labels \cite{mirza_2014_conditional}. The loss function for the cGAN, $L_{c}(D,G)$, can be represented as follows: 
\begin{equation}\begin{split}
     L_{c}(D,G) & = E_{y\sim P_{data}(y)}[\log D(y|x)]\\ & + E_{z\sim P_{z}(z)}[\log (1-D(G(z|x)))].\end{split}\label{eq:3}
\end{equation} Similar to (2), the objective function of the cGAN is minimized in a two-player minimax manner, which is denoted as $L_{cGAN}(D,G,y,x)$ and defined by: \begin{equation}\begin{split}
L_{cGAN}(D,G,y,x) & = \min_{G}\max_{D} [E_{y\sim P_{data}(y)}[\log D(y|x)]\\ & + E_{z\sim P_{z}(z)}[\log (1-D(G(z|x)))]].\end{split}\label{eq:4}
\end{equation}

\section{Proposed Method}

Our proposed method is inspired to further advance cross-spectral iris matching systems utilizing the capabilities of the GAN based approaches. Therefore, we do not generate a synthesized NIR image of its VIS counterpart before matching. Instead, we specifically focus on projecting both the NIR and VIS iris images to a common latent low-dimensional embedding subspace using a generative network. We explore this low-dimensional latent feature subspace for matching iris images in cross-spectral domain with the help of an adversarial network due to its great success in finding optimal solution for synthetic image generation.

\subsection{Deep Adversarial Coupled Framework}

Our proposed conditional CpGAN for iris matching in cross-spectral domain consists of two conditonal GAN (cGAN) modules as shown in Fig. 2. One of them is dedicated to reconstructing the VIS iris images and hence, we refer to as the VIS cGAN module. Similarly, the other module is dedicated to synthesizing the NIR iris images, which is referred to as the NIR cGAN module. In this work, we use a U-Net architecture for the generator to achieve the low-dimensional embedded subspace for cross-spectral iris matching via a contrastive loss along with the standard adversarial loss. In addition to the adversarial loss and contrastive loss \cite{Chopra2005LearningAS}, the perceptual loss \cite{Johnson2016PerceptualLF} and $L_{2}$ reconstruction loss are also used to guide the generators towards the optimal solutions. Perceptual loss is measured via a pre-trained VGG 16 network, which helps in sharp and realistic reconstruction of the images. 

Our prime goal is to match a VIS iris probe against a gallery of NIR iris images, which have not been seen by the network during the training. To perform this matching in the cross-spectral domain, a discriminative model is required to produce a domain invariant representation. Therefore, we focus on learning iris feature representations in a common embedding subspace by incorporating a U-Net auto-encoder architecture that uses class-specific contrastive loss to match the iris patterns in the latent domain.

As previously mentioned, we use a U-Net auto-encoder architecture in our generator for its structural ability of extracting features in the latent embedding subspace. More specifically, the contracting path of the “U shaped” structure of the U-Net captures contextual information, which is passed directly across all the layers, including the bottleneck. Also, the high-dimensional features of the contracting path of the U-Net, combined with the corresponding upsampled features of the symmetric expanding, path provides a means to share the useful information throughout the network. Moreover, during domain transformation, a significant amount of low-level information needs to be shared between input and output, which can be accomplished by leveraging a U-Net-like architecture.

We have followed the architecture of patch-based discriminators \cite{Isola_2017_ImagetoImageTW} to design the discriminators of our proposed model. The discriminators are trained simultaneously along with the respective generators. It is worthwhile to mention that the $L_{1}$ loss performs very well when applied to preserve the low-frequency details but fails to preserve the high-frequency information, whereas patch-based discriminator ensures the preservation of high-frequency details since it penalizes the structure at the scale of the patches.

Although the VIS and NIR iris images are in different domains, they gradually build a connection in the common embedding feature subspace. The features are domain invariant in the embedded subspace, which provides it credibility to discriminate images based on identity. Our final objective is to find a set of domain invariant features in a common embedding subspace by coupling the two generators via a contrastive loss function, $L_{cont}$ \cite{Chopra2005LearningAS}.

The contrastive loss function,  $L_{cont}$, is defined as a distance-based loss metric, which is computed over a set of pairs in the common embedding subspace such that images belonging to the same identity (genuine pairs i.e., a VIS iris image of a subject with its corresponding NIR iris image) are embedded as close as possible, and images of different identities (imposter pairs i.e., a VIS iris image of a subject with a NIR iris image of a different subject) are pushed further apart from each other. The contrastive loss function is formulated as:
\begin{equation}
\begin{split}
L_{cont}(z_1&(x^i_{VIS}),z_2(x^j_{NIR}),Y)= \\ & 
  (1-Y)\frac{1}{2}(D_z)^2 + (Y)\frac{1}{2}(\mbox{max}(0,m-D_z))^2,  
  \end{split}
  \end{equation}where
$x^i_{VIS}$ and $x^j_{NIR}$ denote the input VIS and NIR iris images, respectively. The variable, $Y$, is a binary label, which is set to 0 if $x^i_{VIS}$ and $x^j_{NIR}$ belong to the same class (i.e., genuine pair), and equal to 1 if $x^i_{VIS}$ and $x^j_{NIR}$ belong to different classes (i.e., impostor pair). $z_1(.)$ and $z_2(.)$ are denoted as the encoding functions of the U-Net auto-encoder, which transform  both $x^i_{VIS}$ and $x^j_{NIR}$, respectively into a common latent embedding subspace. Here, $m$, is used as the contrastive margin to ``tighten" the constraint. The Euclidean distance, $D_z$, between the outputs of the functions, $z_1(x^i_{VIS})$, and $z_2(x^j_{NIR})$, is given by:
 
\begin{equation}
     D_z=\left\lVert z_1(x^i_{VIS})-z_2(x^j_{NIR})\right\rVert_2.
\end{equation}
 
 Therefore, if $Y=0$ (i.e., genuine pair), then the contrastive loss function, $(L_{cont})$, is given as:
 \begin{equation}
{L_{cont}(z_1(x^i_{VIS}),z_2(x^j_{NIR}),Y)  = \frac{1}{2}\left\lVert z_1(x^i_{VIS})-z_2(x^j_{NIR})\right\rVert^2_2}, 
\end{equation} and if $Y=1$ (i.e., impostor pair), then contrastive loss function, $(L_{cont})$, is :
  \begin{equation}
  \begin{split}
L_{cont}(z_1(x^i_{VIS}),&z_2(x^j_{NIR}),Y)  = \\ & \frac{1}{2}\mbox{max}\biggl(0,m-\left\lVert z_1(x^i_{VIS})-z_2(x^j_{NIR})\right\rVert^2_2\biggr).
\end{split}
\end{equation}

Thus, the total loss  for coupling the VIS generator and NIR generator is denoted by $L_{cpl}$ and is given as:
\vspace{-0.25cm}
\begin{equation}
    \begin{split}
        L_{cpl}=\frac{1}{N^2}\sum_{i=1}^{N}\sum_{j=1}^{N}L_{cont}(z_1(x^i_{VIS}),z_2(x^j_{NIR}),Y), 
    \end{split}\label{eq:6}
\end{equation}
where N is the number of training samples. The contrastive loss in the above equation can also be replaced by some other distance-based metric, such as the Euclidean distance. However, the main aim of using the contrastive loss is to be able to use the class labels implicitly and find a discriminative embedding subspace, which may not be the case with some other metric such as the Euclidean distance. This discriminative embedding subspace would be useful for matching the VIS iris images against the gallery of NIR iris images.

\section{Loss Functions}

\subsection{Generative Adversarial Loss}

Here, we denote $G_{VIS}$ and $G_{NIR}$ as the VIS and NIR generators that reconstruct the corresponding VIS and NIR iris images from the input VIS and NIR iris images, respectively. $D_{VIS}$ and $D_{NIR}$ are denoted as the patch-based discriminators used for the VIS and NIR iris GANs. Since we have used a conditional GAN for our proposed method, we condition both the generator networks, $G_{VIS}$ and $G_{NIR}$, on input VIS and NIR iris images, respectively. In addition, we have trained the generators and the corresponding discriminators with the conditional GAN loss function \cite{mirza_2014_conditional} to ensure the reconstruction of real-looking natural image such that the discriminators cannot distinguish the generated images from the real ones.  Let $L_{VIS}$ and $L_{NIR}$ denote the conditional GAN loss functions for the VIS and NIR GANs, respectively, where $L_{VIS}$ and $L_{NIR}$ are given as: 
\vspace{-0.2cm}
\begin{equation}
L_{VIS}=L_{cGAN}(D_{VIS},G_{VIS},y^i_{VIS},x^i_{VIS}),
\end{equation}
\begin{equation}
L_{NIR}=L_{cGAN}(D_{NIR},G_{NIR},y^j_{NIR},x^j_{NIR}), 
\end{equation}where $L_{cGAN}$ is defined as the conditional GAN objective function in (\ref{eq:4}). The term, $x^i_{VIS}$, is used to denote the VIS iris image, which is defined as a condition for the VIS GAN, and $y^i_{VIS}$,  is denoted as the real VIS iris image. It is worth mentioning that the real VIS iris image, $y^i_{VIS}$, is same as the network condition given by $x^i_{VIS}$. Similarly, $x^j_{NIR}$, denotes the NIR iris image that is used as a condition for the NIR GAN. Again, like  $y^i_{VIS}$, the real NIR iris image, $y^j_{NIR}$, is same as the network condition given by $x^j_{NIR}$ . The total objective function for the coupled conditional GAN is given by: 

\begin{equation}
L_{GAN}=L_{VIS}+L_{NIR}.
\end{equation}

\subsection{$L_2$ Reconstruction Loss}

For both the VIS GAN and NIR GANs, we consider the $L_2$ reconstruction loss as a classical constraint to ensure better results. The $L_2$ reconstruction loss is measured in terms of the Euclidean distance between the reconstructed iris image and the corresponding real iris image.
We denote the reconstruction loss for the VIS GAN as $L_{2_{VIS}}$ and define it as follows:

\begin{equation}
    L_{2_{VIS}}=\left\lVert G_{VIS}(z|x^i_{VIS})-y^i_{VIS}\right\rVert^2_2,
\end{equation}where $y^i_{VIS}$ is the ground truth VIS iris image, and $G_{VIS}(z|x^i_{VIS})$, is the output of the VIS generator.

Similarly, lets denote the reconstruction loss for the NIR GAN as $L_{2_{NIR}}$: 

\begin{equation}
    L_{2_{NIR}}=\left\lVert G_{NIR}(z|x^j_{NIR})-y^j_{NIR}\right\rVert^2_2,
\end{equation}where $y^j_{NIR}$ is the ground truth NIR iris image, and $G_{NIR}(z|x^j_{NIR})$, is the output of the NIR generator.

The total $L_2$ reconstruction loss can be given by the following equation:
\begin{equation}
    L_{2}=\frac{1}{N^2}\sum_{i=1}^{N}\sum_{j=1}^{N}(L_{2_{VIS}}+L_{2_{NIR}}).
\end{equation}

\subsection{Perceptual Loss}\label{subsec:percloss}
Although the GAN loss and the reconstruction loss are used to guide the generators, they fail to reconstruct perceptually pleasing images. Perceptually pleasing means images with perceptual features defined by the visual deterministic properties of objects. Hence, we have also used perceptual loss introduced in \cite{Johnson2016PerceptualLF} for style transfer and super-resolution. The perceptual loss function basically measures high level differences, such as content and style dissimilarity, between images. The perceptual loss is based on high-level representations from a pre-trained VGG-16 \cite{simonyan2014very} like CNN. Moreover, it helps the network generate better and sharper high quality images \cite{Johnson2016PerceptualLF}. As a result, it can be a good alternative to solely using  $L_1$ or $L_2$ reconstruction error. 

In our proposed approach, we have added perceptual loss to both the VIS and NIR GAN modules using a pre-trained VGG-16 \cite{simonyan2014very} network. It involves extracting the high-level features (ReLU3-3 layer) of VGG-16 for both the real input image and the reconstructed output of the U-Net generator. The perceptual loss calculates the $L_1$ distance between the features of real and reconstructed images to guide the generators $G_{VIS}$ and $G_{NIR}$. The perceptual loss for the VIS GAN network is defined as:

\begin{equation}
    \begin{split}
        L_{P_{VIS}}=&\frac{1}{C_pW_pH_p}\sum_{c=1}^{C_{p}}\sum_{w=1}^{W_{p}}\sum_{h=1}^{H_{p}} \\ & {\left\lVert V(G_{VIS}(z|x^i_{VIS}))^{c,w,h}-V(y^i_{VIS})^{c,w,h}\right\rVert}_2^2,
    \end{split}
\end{equation}
where $V(.)$ is used to denote a particular layer of the VGG-16 and $C_p$, $W_p$, and $H_p$ denote the layer dimensions.

Likewise the perceptual loss for the NIR GAN network is:

\begin{equation}
    \begin{split}
        L_{P_{NIR}}=&\frac{1}{C_pW_pH_p}\sum_{c=1}^{C_{p}}\sum_{w=1}^{W_{p}}\sum_{h=1}^{H_{p}} \\ & {\left\lVert V(G_{NIR}(z|x^j_{NIR}))^{c,w,h}-V(y^j_{NIR})^{c,w,h}\right\rVert}_2^{2}.
    \end{split}
\end{equation}

The total perceptual loss function is given by:
\begin{equation}
    L_{P}=\frac{1}{N^2}\sum_{i=1}^{N}\sum_{j=1}^{N}(L_{P_{VIS}}+L_{P_{NIR}}).
\end{equation}

\subsection{Overall Objective Function}
We sum up all the loss functions defined above to calculate the overall objective function for our proposed method: 
\vspace{-0.25cm}
\begin{equation}
    \begin{split}
       L_{tot}=L_{cpl}+ \lambda_1 L_{GAN} + \lambda_2 L_{P}+ \lambda_3 L_2,
    \end{split}\label{eq:20}
\end{equation}where $L_{cpl}$ is the coupling loss, $L_{GAN}$ is the total generative adversarial loss, $L_{P}$ is the total perceptual loss, and $L_2$ is the total reconstruction error. Variables $\lambda_1$, $\lambda_2$, and $\lambda_3$ are the hyper-parameters used as a weight factor to numerically balance the magnitude of the different loss terms. 
\vspace{0.5cm}
\section{Experiments}
\vspace{-0.1cm}
In this section, we first describe the datasets and the training details to show the implementation of our method. To show the efficiency of our method for the task of iris recognition in cross domain, we compare its performance with other existing cross-spectral iris recognition methods. 
\vspace{-0.17cm}


\subsection{Experimental Details}
\textbf{Datasets}: 
We conduct the experiments using the PolyU Bi-spectral database. The PolyU Bi-Spectral database \cite{nalla2016toward, wang2019cross} (see Figure 1) contains iris images of 209 subjects obtained simultaneously in both the VIS and NIR wavelengths. The data for each subject consists of 15 different instances of right-eye images and left-eye images for both VIS and NIR spectrum. Therefore, the total number of images in this dataset is 12,540 with a resolution of $640 \times 480$ pixels. For the experiment, we split the dataset into training and testing sets. We choose the images of the last 168 of the identities as the training set and all the images of the remaining identities as the testing set. 

\textbf{Implementation Details}: 
We have implemented our CpGAN architecture using the U-Net architecture as the generator module. We follow the typical CNN architecture for the implementation of both encoder and decoder sections of the U-Net model. The encoder section is designed by applying two $3\times3$ convolutions, each followed by a rectified linear unit (ReLU). For downsampling, it uses $2\times2$ max pooling operation with stride 2. We double the number of feature channels at each downsampling step. Similarly, each step in the decoder section upscales the feature map by applying a $2\times2$ transpose convolution convolution (“deconvolution that is similar to upconvolution”) and halves the number of feature channels. After upsampling the dimension of the feature map, each feature map is concatenated with the corresponding feature map from the encoder, followed by two $3\times3$ convolutions with a ReLU activation function. 


The proposed framework has been implemented in Pytorch. We trained the network with a batch size of 16 and a learning rate of 0.0002. We used the Adam optimizer \cite{kingma2019method} with a first-order momentum of 0.5, and a second-order momentum of 0.999. We have used the Leaky ReLU as the activation function with a slope of 0.35 for the discriminator. For the network convergence, we set 1 for $\lambda_1$, and 0.3 for both $\lambda_2$, and $\lambda_3$. 

For training, genuine/impostor pairs are created from the VIS and NIR iris images of the same/different subjects. During the experiments, we ensure that  the training set is balanced by using the same number of genuine and impostor pairs.

\section{Evaluation on PolyU Bi-Spectral Database}

We have evaluated our proposed method on the PolyU Bi-Spectral benchmark iris dataset. The PolyU Bi-Spectral iris dataset contains co-registered eye images in VIS as well as NIR spectrum. We conduct several experiments to show the efficacy of our proposed scheme. In all experiments, each probe image of the test set is matched against a gallery of images which are in a different domain (e.g., VIS or NIR). As a consequence, we obtain genuine and imposter scores, which guide calculation of the essential recognition performance parameters, such as genuine acceptance rate (GAR), false acceptance rate (FAR), and equal error rate (EER). In addition, we plot receiver operating characteristics (ROC) curves to analyze the GAR with respect to FAR.
We have studied the following cases for cross-spectral iris matching:

\hspace{-0.4cm}(a) \textbf{Matching High Resolution VIS iris images against a gallery of High Resolution NIR iris images}

\vspace{0.2cm}

In this experiment, we train our network with the unrolled high resolution $(64\times 512)$ VIS and NIR iris images such that the VIS and NIR generators are trained to obtain domain invariant features in a common embedding subspace. 

Our purpose is to use the trained network for matching high resolution (HR) VIS iris images against a gallery of high resolution (HR) NIR iris images, which were unseen by the network during the training. We evaluate the performance of this network on the PolyU Bi-Spectral dataset. To show the comparative performances, we consider other state-of-the-art deep learning approaches (Wang et al. \cite{wang2019toward,wang2019cross}, and Oktiana et al. \cite{oktiana2019advances}), which apply different types of feature extraction techniques. In addition, we have plotted ROC curves comparing our proposed approach with the baseline algorithms already mentioned above. The results are summarized in Table \rom{1}.

From Fig. 3 and Table \rom{1}, we observe that our proposed CpGAN framework performs much better than the other baseline (Wang et al. \cite{wang2019cross,wang2019toward} and Oktiana et al. \cite{oktiana2019advances}) matching algorithms. In this setting, our method achieves 1.67\% more identification accuracy with 4.37\% decrease in EER compared to the most recent cross-spectral iris recognition method \cite{wang2019cross}. Additionally, it outperforms the method described in \cite{wang2019toward,oktiana2019advances} by a significant decrease of 0.67\% and 16.01\% in EER, respectively. This significant improvement clearly indicates that the usage of a CpGAN framework for projecting the VIS and NIR iris images into the latent embedding subspace to retrieve the domain invariant features is better than the other existing deep learning methods. 
\hspace{-0.4cm}(b) \textbf{Matching High Resolution VIS iris images against a gallery of Low Resolution NIR iris images}

Here, we analyze our network by considering an ideal scenario for cross-spectral iris recognition systems. We mentioned that in surveillance-based iris recognition systems, surveillance cameras capture high-resolution images under visible spectrum, while images already stored in the gallery are in the NIR domain and having a lower resolution. Therefore, it has become a challenging issue for the existing cross-spectral iris recognition systems to ascertain the correlation between iris images in different resolutions as well as at different spectra. Their limitations of retrieving accurate semantic similarity in the iris images of different resolution and spectrum have resulted in a significant performance degradation. One of the most probable ways to resolve this issue could be the usage of CpGAN network trained with the unrolled high-resolution (HR) ($64\times512$) VIS and low-resolution (LR) ($32 \times 256$) NIR iris images, which ensures the retrieval of contextual and semantic features of the iris images in a common embedding subspace. To verify the usefulness of this network, we perform matching HR VIS iris images against a gallery of LR NIR iris images using the publicly available PolyU Bi-spectral dataset. The results summarized in Fig. 3 and Table \rom{1} indicate that the network remains robust enough to provide outperforming results compared to the methods described in \cite{oktiana2019advances,wang2019toward}, which may have significant contribution to the real-life applications. 

\begin{figure}[t]
\centering
\includegraphics[width=7.5cm]{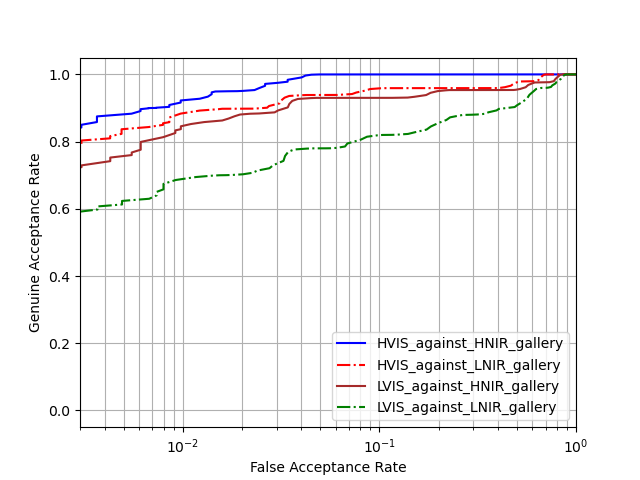}
\caption{ROC curves showing the results obtained on the PolyU Bi-Spectral database.}
\label{fig:arch}
\end{figure}

\begin{table}[t]
\centering
\caption{Comparative performances on the PolyU Bi-Spectral database. Symbol '-' indicates that the metric is not available for that protocol.}
\vspace{0.4cm}
\scalebox{0.62}{\begin{tabular}{c c c c c}
 \hline
Algorithm&Matching&GAR@FAR=0.01&GAR@FAR=0.001&EER  \\ \hline
Wang et al. \cite{wang2019toward}&HR VIS vs HR NIR&59.10&37.00&17.03 \\ \hline
CNN with SDH \cite{wang2019cross} &HR VIS vs HR NIR&90.71&84.50&5.39 \\ \hline
Garg et. al \cite{garg2018heterogeneous}& HR VIS vs HR NIR&|&48.86\\ \hline
GRF BSIF \cite{oktiana2019advances} &HR VIS vs HR NIR&82.92&79.12&1.69 \\ \hline
GRF LBP \cite{oktiana2019advances} &HR VIS vs HR NIR&69.23&67.04&4.2 \\ \hline
\textbf{Ours (CpGAN)} &\textbf{HR VIS vs HR NIR}&\textbf{92.38}&\textbf{84.98}&\textbf{1.02} \\ \hline
{Ours (CpGAN)}&{HR VIS vs LR NIR}&{89.89}&{81.21}&{1.21} \\ \hline
Ours (CpGAN)&HR NIR vs LR VIS&84.75&73.45&1.26 \\ \hline
Ours (CpGAN)&LR NIR vs LR VIS&70.10 & 59.97&2.51 \\ \hline
\end{tabular}}
\label{table:table_cfp}
\end{table}
\hspace{-0.4cm}(c) \textbf{Matching Low Resolution VIS iris images against a gallery of High Resolution NIR iris images}

In most cross-spectral iris recognition systems, researchers have focused on matching high- resolution VIS iris images against a gallery of low-resolution NIR iris images. They did not consider to the scenario where matching could be performed using low-resolution VIS probe iris images. To illustrate the point, it is worth considering that surveillance cameras in public areas having a large field of view often capture images of the subjects which are at a large standoff distance from the camera \cite{garg2018heterogeneous}. Due to this fact, captured faces are expected to be in low-resolution, which suffer from poor quality. On the other hand, the gallery images have high-resolution which are generally collected in the NIR spectrum. So, matching with such a modality gap between probe and gallery images makes the cross-spectral recognition problem even more challenging. Hence, we emphasize this surveillance scenario and train the VIS and NIR generator of our network with the unrolled LR VIS iris images ($32\times 256$) and HR NIR iris images ($64\times512$), respectively. The matching is performed in the latent embedded subspace, as it contains all the information about the iris texture patterns irrespective of the resolution.

From Fig. 3 and Table 1, we observe that our proposed algorithm achieves 84.75\% genuine acceptance rate (GAR) at 0.01 FAR and 1.31\% equal error rate (EER), which, as in the previous test cases, outperforms the results reported by Oktiana et al. \cite{oktiana2019advances} and Wang et al. \cite{wang2019toward}. The network obtains 4.13\% and 15.77\% less EER compared to the results reported in \cite{wang2019toward, wang2019cross}, respectively. 

\hspace{-0.4cm}(d) \textbf{Matching Low Resolution VIS iris images against a gallery of Low Resolution NIR iris images}

In addition to the study mentioned above, we have also investigated the matching performance of our network when our gallery images are in a low-resolution NIR domain. 

To train our network, we feed both the VIS and NIR generator with the unrolled LR VIS and NIR iris images. The experimental results indicate the matching accuracy of this network, which are reported in Table \rom{1} and Fig. 3. It is observed that, even though it achieves an EER of 2.51\% that is much lower than  several comparable methods, the verification performance is not as satisfactory as our previous experiments outlined above.

\hspace{-0.4cm}(e) \textbf{Cross-Spectral Iris Matching in Synthesized Domain}

In order to achieve accurate iris recognition performance in the cross-spectral domain, many researchers applied domain transformation techniques before matching. However, in our proposed method we perform matching in a modality-invariant embedded subspace utilizing its latent feature vectors. A range of experiments has been conducted to validate the efficacy of our proposed method. Moreover, we have also investigated the impact of the domain transformation technique followed by the cross-spectral iris matching. Several experiments have been performed to quantify the performance of our model regarding these following scenarios:

\hspace{-0.4cm}(1) We have utilized the feature vectors generated in the embedding subspace of the VIS cGAN generator to produce synthesized NIR iris images. In more detail, the VIS cGAN generator is trained with the unrolled HR VIS iris images such that the generated feature vectors can be fed to the decoder section of the NIR cGAN generator network to synthesize the corresponding HR NIR iris images. These synthesized HR NIR iris images are then matched against the HR NIR iris gallery. We used the OSIRIS \cite{othman2016osiris} software for matching. \vspace{0.1cm}

\hspace{-0.4cm}(2) Similarly, the feature vectors generated from the NIR cGAN generator are used as input to the decoder of the VIS cGAN network to obtain synthesized HR VIS iris images which are matched against the HR VIS iris gallery.\vspace{0.1cm}

\hspace{-0.4cm}(3) To show the iris matching performance in cross-spectral as well as cross-resolution domains, we conduct additional experiments. More specifically, we train both the VIS and NIR cGAN networks with the unrolled low-resolution VIS and NIR iris images, respectively, so that the representative feature vectors generated in the latent embedding subspace can be employed to reconstruct HR synthesized iris images in the cross-spectral domain following the above manner.

\begin{figure}[t]
\centering
\hspace{-0.35cm}
 \includegraphics[width=9cm]{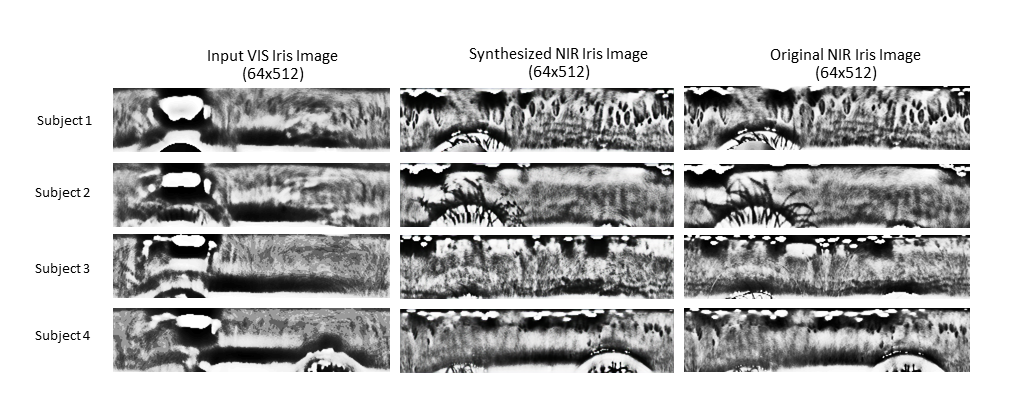}
 \caption{Reconstruction of synthesized HR NIR iris images from the output of the NIR cGAN generator as the HR VIS iris input to the VIS cGAN generator.}
 \label{fig:profile_frontal_images}
\end{figure}
\begin{figure}[t]
\centering
  \includegraphics[width=9cm]{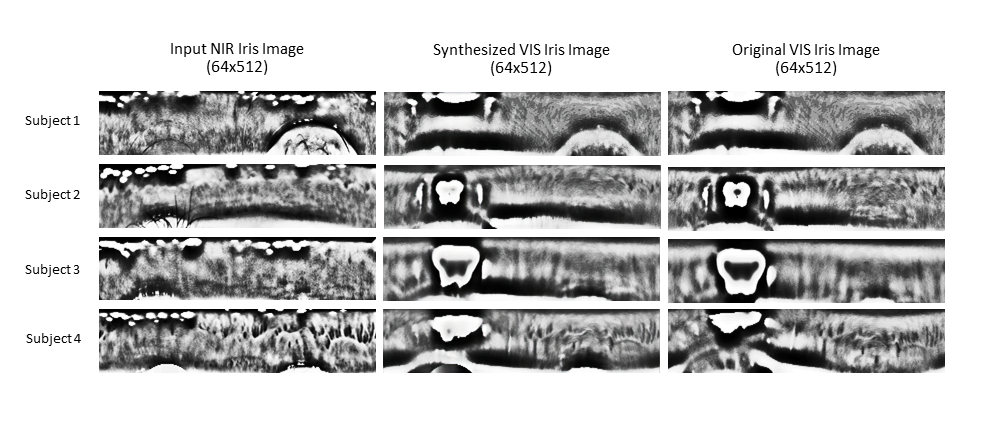}
 \caption{Reconstruction of synthesized HR VIS iris images from the output of the VIS cGAN generator as the HR NIR iris input to the NIR cGAN generator.}
 \label{fig:profile_frontal_images}
\end{figure}

\begin{figure}[t]
\centering
 \includegraphics[width=7.5cm]{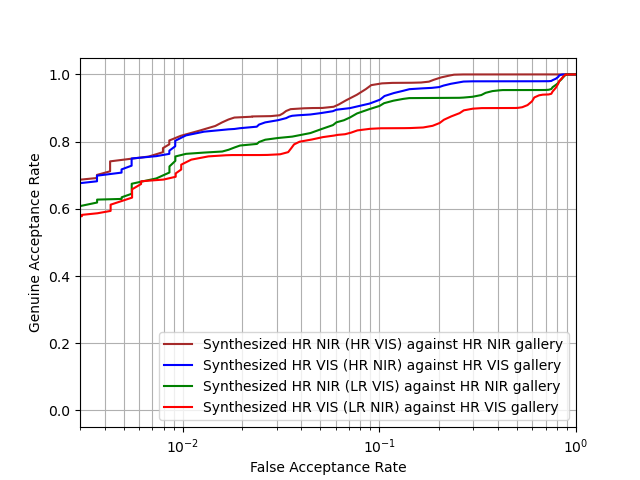}
 \caption{ROC curves showing the performance of our proposed conditional CpGAN network for iris matching in synthesized domain on the PolyU Bi-Spectral iris database.}
 \label{fig:profile_frontal_images}
\end{figure}

\begin{table}[t]
\centering
\caption{Results summary of our proposed conditional CpGAN network for iris matching in synthesized domain on the PolyU Bi-Spectral database.The term in (' ') indicates the input domain.}
\vspace{0.2cm}
\scalebox{0.66}{\begin{tabular}{c c c c c}
 \hline
Algorithm&Matching&GAR@FAR=0.01&EER  \\ \hline
Ours (CpGAN) &Synthesized HR NIR (HR VIS)-HR NIR&83.15&1.12\\ \hline
Ours (CpGAN)&Synthesized HR VIS (HR NIR)-HR VIS&82.16&1.19\\ \hline
Ours (CpGAN) &Synthesized HR NIR (LR VIS)-HR NIR&77.31&1.47\\ \hline
Ours (CpGAN)&Synthesized HR VIS (LR NIR)-HR VIS&73.51&2.72 \\ \hline
\end{tabular}}
\label{table:table_cfp}
\vspace{-0.4cm}
\end{table}

We  used OSIRIS software to perform matching for the set of experiments described above. The ROC curves from this set of experiments are shown in Fig. 6 while verification performance with EER results are summarized in Table 2. We have also shown the synthesized HR results in Fig. 4 and 5. The experimental results indicate that following the domain transformation technique to conduct matching in the same domain where the gallery images belong does not offer as much improvement as our approach, which conducts matching in the latent
embedding subspace.

However, one of the scenarios still outperforms the other baseline methods \cite{oktiana2019advances, wang2019toward} such as HR VIS input to synthesized HR NIR  matched against HR NIR gallery by 0.23\% and 24.05\% GAR at 0.01 FAR, respectively. 
\vspace{-0.2cm}
\section{Conclusion}
\vspace{-0.2cm}
In this paper, we have investigated the cross-domain iris recognition problem and introduced a new approach for more accurate cross-spectral iris matching. We developed a conditional Coupled GAN (CpGAN) framework which projects modality invariant iris texture features in the latent embedding subspace to perform matching in the embedded domain. Matching results outperforming other methods reported in the literature, illustrated in Section \rom{7} of this paper on publicly available PolyU cross-spectral iris database, validate the superiority and effectiveness of our approach.

{\small
\bibliographystyle{ieee}
\bibliography{submission_example}
}

\end{document}